%% file: xxxmain.tex
\documentclass[letterpaper,twocolumn,10pt]{article}
\usepackage{usenix-2020-09}

\makeatletter
\def\@maketitle{\newpage
 \vbox to 1.75in{  
 \vspace*{0.2in}  
 \vskip 2em
 \begin{center}%
  {\Large\bf \@title \par}%
  \vskip 0.375in minus 0.300in
  {\large\it
   \lineskip .5em
   \begin{tabular}[t]{c}\@author
   \end{tabular}\par}%
 \end{center}%
 \par
 \vspace*{0.2in}  
 }
}
\makeatother

\definecolor{codegreen}{rgb}{0,0.6,0}
\definecolor{codegray}{rgb}{0.5,0.5,0.5}
\definecolor{codepurple}{rgb}{0.58,0,0.82}
\definecolor{backcolour}{rgb}{0.95,0.95,0.92}
\definecolor{ballblue}{rgb}{0.13, 0.67, 0.8}
\definecolor{blue}{rgb}{0.0, 0.53, 0.74}
\definecolor{cobalt}{rgb}{0.0, 0.28, 0.67}
\definecolor{coolblack}{rgb}{0.0, 0.18, 0.39}
\definecolor{darkcerulean}{rgb}{0.03, 0.27, 0.49}

\newcommand*\circleo[1]{\tikz[baseline=(char.base)]{
            \node[shape=circle,draw=orange,inner sep=0.5pt,fill=orange,text=white, scale=1] (char) {#1};}}

\newcommand*\circlep[1]{\tikz[baseline=(char.base)]{
            \node[shape=circle,draw=violet!85,inner sep=0.5pt,fill=violet!85,text=white, scale=0.85] (char) {#1};}}

\newcommand*\circled[1]{\tikz[baseline=(char.base)]{
            \node[shape=circle,draw=darkcerulean,inner sep=0.5pt,fill=darkcerulean,text=white, scale=0.85] (char) {#1};}}

\usepackage{tikz}
\usepackage{amsmath}
\usepackage{enumitem}

\usepackage{graphicx}
\graphicspath{ {./images/} }

\usepackage{filecontents}
\usepackage[ruled,vlined,linesnumbered]{algorithm2e}

\newcommand{\xuehai}[1]{\textcolor{black}{#1}}
\newcommand{\yikang}[1]{\textcolor{black}{#1}}
\newcommand{\yishu}[1]{\textcolor{black}{#1}}
\usepackage{soul}

\begin{document}

\date{}


\title{\Large\bf
GreedySnake: Accelerating SSD-Offloaded LLM Training with\\
Efficient Scheduling and Optimizer Step Overlapping
}

\author{
  \textbf{\rm Yishu Yin} \ \ \ \
  \textbf{\rm Xuehai Qian}\thanks{Corresponding author: Xuehai Qian.} \\
  Tsinghua University
}

\maketitle

\input{abstract}
\input{introduction}

\input{background}
\input{analysis}
\input{design}
\input{implementation}
\input{evaluation}


\section{Conclusion}
This paper proposes GreedySnake, a novel SSD-offloaded training system that adopts \emph{vertical scheduling}, where all micro-batches of a layer are executed before advancing to the next layer.
To further alleviate the I/O bottleneck, GreedySnake overlaps part of the optimizer step with the forward pass of the next iteration.
GreedySnake achieves higher throughput with smaller batch sizes.
Experiments on NVIDIA A100 GPUs show that GreedySnake improves saturated training throughput over ZeRO-Infinity by $1.96\times$ (1 GPU) and $1.93\times$ (4 GPUs) for GPT-65B, and by $2.53\times$ (1 GPU) for GPT-175B.

\bibliographystyle{plain}

\end{document}

%% file: abstract.tex
\begin{abstract}


SSD-offloaded training offers a practical and promising
approach to making LLM training cost-effective.
Building on gradient accumulation with micro-batches,
this paper introduces GreedySnake, a new SSD-offloaded training
system that employs \emph{vertical scheduling},
which executes all micro-batches of a layer before proceeding to the next.
Compared to existing systems that use \emph{horizontal scheduling}
(i.e., executing micro-batches sequentially),
GreedySnake achieves higher training throughput with smaller batch sizes,
bringing the system much closer to the ideal scenario predicted by the roofline model.
To further mitigate the I/O bottleneck,
GreedySnake overlaps part of the optimization step with the forward pass of the next iteration.
\xuehai{Experimental results on A100 GPUs show that GreedySnake achieves saturated training throughput improvements over ZeRO‑Infinity: 1.96× on 1 GPU and 1.93× on 4 GPUs for GPT‑65B, and 2.53× on 1 GPU for GPT‑175B.}

\end{abstract}

%% file: introduction.tex
\section{Introduction}

Large Language Models (LLMs) have attracted widespread attention across diverse domains~\cite{vaswani2017attention,Brown2020LanguageMA,Chowdhery2022PaLMSL}.
Scaling laws~\cite{Kaplan2020ScalingLF} show that LLM performance improves with increased model parameters, training data, and compute.
Motivated by this trend, recent open-source LLMs such as gpt-oss~\cite{Agarwal2025gptoss120bgptoss20bMC}, Qwen~\cite{Bai2023QwenTR}, and DeepSeek~\cite{deepseekai2025deepseekv3technicalreport} often exceed 100 billion parameters.
To train such large models cost-effectively, heterogeneous memory training has been proposed, where most of the training footprint resides outside GPU memory.
Early systems~\cite{rhu2016vdnn,wang2018superneurons,huang2020swapadvisor,hildebrand2020autotm,peng2020capuchin,ren2021zerooffload,ren2021sentinel,fang2022patrickstar,sun2022stronghold,lian2025superoffload} leverage CPU memory, and later designs, known as \emph{SSD-offloaded} training~\cite{rajbhandari2021infinity,wu2025ssdtrainactivationoffloadingframework,liao2025ratel,yuan2025cost}, go a step further by storing various types of training data across both CPU memory and SSD.
Without relying solely on GPU memory to host all training data, this approach offers a promising solution for LLM training under GPU memory-constrained settings.
In this paper, we rethink the fundamental challenges of SSD-offloaded training and propose novel techniques that significantly improve its efficiency compared to state-of-the-art systems.

\yishu{SSD-offloaded} training typically offloads the large optimizer states to SSD and uses the host CPU to execute the optimizer step and update these states~\cite{rajbhandari2021infinity,liao2025ratel}.
In this setting, the optimizer step becomes the primary bottleneck due to excessive I/O: optimizer states must be loaded from SSD for updates and written back afterward, saturating the limited host-SSD bandwidth (typically a few GB/s).
Jang \emph{et al.}~\cite{jang2024smart} report that this overhead can reach 80\% when training GPT-2.

To facilitate data movement and reduce storage access time, prior works~\cite{liao2025ratel,yuan2025cost} suggest deploying up to 12 SSDs or using two 4-SSD RAID0 arrays to increase host-storage bandwidth, or leveraging in-storage computing~\cite{jang2024smart} to reduce storage traffic.
However, these approaches require additional or specialized hardware.
In practice, conventional servers are rarely provisioned with so many SSDs to scale bandwidth, making these solutions unlikely to see widespread adoption.


\begin{figure*}[htbp]
    \centering
    \includegraphics[width=\linewidth]{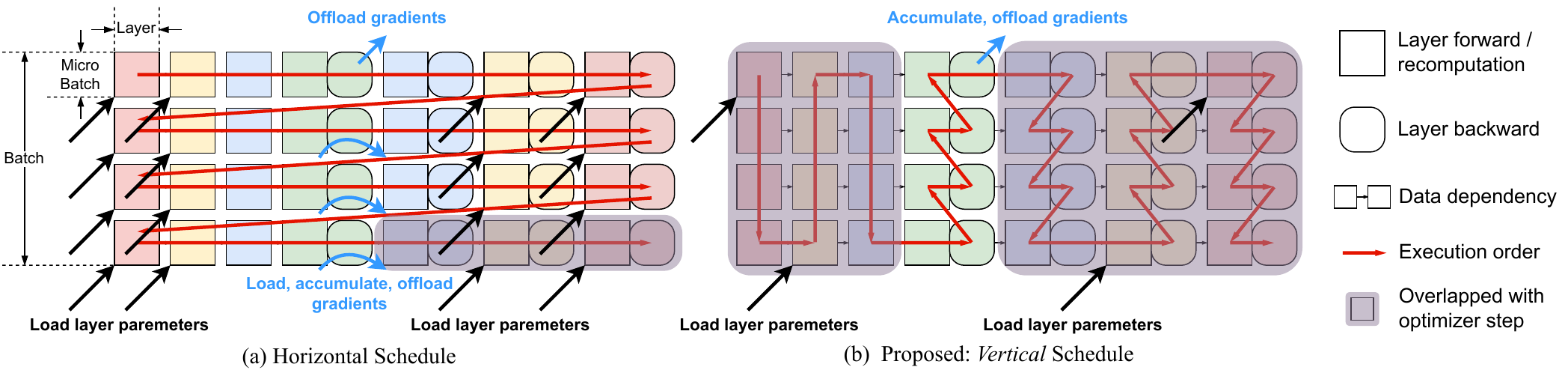}
  \caption{Gradient Accumulation: From Horizontal to Vertical. The swapping traffic is only plotted for a subset of layers. }
    
  \label{fig:idea}
  \label{fig:overview}
\end{figure*}

To tackle the optimizer step bottleneck, a common practice is to use larger batch sizes so that the optimizer step overhead---independent of batch size---can be amortized over an entire training iteration.
Some prior works~\cite{bae2021flashneuron,fang2022patrickstar,liao2025ratel} focus on increasing the batch size within a single forward-backward pass.
However, the maximum batch size is ultimately capped by the peak memory usage of the largest operator in the model, e.g., matrix multiplications in feed-forward networks.
More importantly, increasing batch size causes faster growth in the data traffic for transferring \emph{activation checkpoints} required by recomputation~\cite{chen2016checkpointing}, a widely used memory-saving technique.
This technique creates activation checkpoints periodically---typically at layer boundaries---during the forward pass and reconstructs intermediate activations between checkpoints in the backward pass.
A larger batch size increases not only the size of each checkpoint, but also the \emph{frequency} of checkpoints, because the amount of recovered activations between checkpoints is bounded by GPU memory capacity.
These larger and more frequent checkpoints lead to \emph{superlinear} growth in the overhead of checkpoint swapping, quickly wiping out the benefits of increasing batch size.

A fundamentally better approach to scaling batch size is
{\em gradient accumulation} adopted by
ZeRO-Infinity~\cite{rajbhandari2021infinity}, which computes and aggregates gradients from multiple {\em micro-batches} within each iteration, scaling the 
effective batch size by increasing the number of micro-batches.
Specifically, consecutive micro-batches of an iteration
are executed one by one, and the gradients produced by each micro-batch are accumulated. 
However, gradient accumulation comes with
its own overhead. 
We analyze the problem from two aspects that directly affect 
training performance:
(1) the amount of data movement; and 
(2) the ability of overlapping optimizer steps with GPU computation.

Consider an $N$-layer LLM with total model size $ms$. Each training iteration processes $M$ micro-batches under gradient accumulation, where the aggregated activation checkpoint size per micro-batch is $cs$.
Current systems perform gradient accumulation with a \emph{horizontal schedule} that executes micro-batches sequentially.
This horizontal schedule incurs data movement between the GPU and the lower memory hierarchy from three sources:
(1) loading model parameters with total size $2 \times M \times ms$, i.e., parameters are loaded for both the forward pass and the backward pass with recomputation for each micro-batch;
(2) writing activation checkpoints in the forward pass and reading them in the backward pass, with total size $2 \times M \times cs$; and
(3) loading, updating, and writing back gradients of all parameters across $(M-1)$ micro-batches, leading to a total of $[2(M-1)+1] \times 2ms = (2M-1) \times 2ms$%
\footnote{We assume mixed-precision training~\cite{micikevicius2018mixed}. Gradient accumulation is typically kept in full precision, so the buffer size is twice the parameter size. The first micro-batch only offloads gradients; the remaining $(M-1)$ micro-batches fetch and offload gradients.}.
In this setting, the optimizer step can be overlapped with the backward pass of the last micro-batch for $(N-1)$ layers.


This paper reexamines the computation schedule for SSD-offloaded training and proposes \emph{GreedySnake}, a system based on the \emph{vertical schedule} that executes each layer's forward or backward computations across \emph{all micro-batches} before advancing to the next layer.
GreedySnake explores a key tradeoff between reduced model-parameter loading and increased activation checkpoint reads and writes.
This elegant yet simple design naturally reduces overall data movement and substantially increases the amount of GPU computation that can be overlapped with the optimizer step.

On one side, the vertical schedule enables better \emph{reuse} of loaded parameters and reduces parameter loading traffic from $2 \times M \times ms$ to $2ms$.
The traffic for moving gradients is similarly reduced to $2ms$: parameter gradients from all micro-batches are accumulated locally in GPU memory, and the fully accumulated gradients for each layer are transferred to CPU memory only \emph{once}.
On the other hand, GreedySnake increases activation checkpoint traffic. Because execution switches across micro-batches at each layer, the intermediate activations of a given micro-batch are no longer kept in GPU memory when its next layer is executed; instead, they must be saved as checkpoints and later reloaded to continue the forward pass. A similar issue arises in the backward pass, where the inter-layer gradients are likewise unavailable in GPU memory when a micro-batch resumes.
The \emph{key insight} of GreedySnake is that parameter size scales \emph{quadratically} with the model hidden dimension, whereas checkpoint size scales \emph{linearly} (see Section~\ref{sec:our-schedule}).
In other words, improving parameter reuse is more critical.

Naturally, vertical scheduling substantially increases the opportunity to overlap the optimizer step with the backward pass.
Specifically, the overlapped computation becomes approximately $M \times (N-1)$ layers of backward passes across \emph{all micro-batches}, up from only $(N-1)$ layers of the backward pass for \emph{the last micro-batch}.
Taking a step further, GreedySnake also overlaps part of the optimizer step with the forward pass of the next iteration, as long as each layer's parameters are updated before that layer executes.
By combining these two techniques, our system can ideally overlap the optimizer step with almost all GPU computation across all micro-batches, providing unprecedented tolerance to long optimizer step execution.
Figure~\ref{fig:overview} compares the horizontal and vertical schedules and illustrates the benefits of the latter.

\xuehai{Following the roofline model of SSD-offloaded training (Figure~\ref{fig:roofline}), as batch size increases, training throughput initially grows under the I/O roofline and then saturates below the computation roofline.
Our goal is to (i) maximize this saturated throughput and (ii) minimize the batch size required to reach it.
Doing so improves algorithmic flexibility: users need not inflate batch size solely due to system constraints.
In GreedySnake, vertical scheduling increases the highest achievable throughput, while aggressive overlap reduces the batch size required to attain it.
However, both the saturation point and the batch size at which it is reached depend on (1) how training data are distributed across GPU memory, CPU memory, and SSD, and (2) the percentage of the optimizer step to be overlapped with the forward pass of the next iteration.
We formulate a linear program to automatically search this large configuration space and identify the best data-distribution and overlap ratios.}

We implement GreedySnake in approximately 5K lines of Python (excluding comments and blank lines) on top of PyTorch~\cite{pytorch} to manage computation and tensor storage, and we reuse the \texttt{asyncio}-based pipeline and the \texttt{cpu\_adam} module from ZeRO-Infinity~\cite{rajbhandari2021infinity}.
Our evaluation compares GreedySnake against ZeRO-Infinity and several other recent SSD-offloaded training systems.
We run experiments on two GPU clusters and train GPT-30B, GPT-65B, and GPT-175B.
The results show that GreedySnake substantially improves the saturated training throughput over all baselines.

%% file: background.tex
\section{Background}


\subsection{LLM Training Preliminaries}
\label{sec:basic}

\textbf{Vanilla LLM training.}
Vanilla LLM training proceeds in three stages: \emph{forward} pass, \emph{backward} pass, and \emph{optimizer step}.
During the forward pass, the input is propagated layer by layer, going through each layer's
parameters to generate activations.
The activations produced by the last layer 
are used to compute the loss, while intermediate activations are retained for gradient computation.
In the backward pass, the loss signal is propagated reversely through all layers, combining with stored activations produced in the forward stage 
to compute gradients for each parameter.
Finally, in the optimizer step, parameters are updated element‑wise using the parameter gradients together with optimizer states, producing updated parameters and optimizer states for the next iteration.

\textbf{Pipelined optimizer step.}
A straightforward optimization for vanilla LLM training is to pipeline the backward and optimizer‑step stages.
\xuehai{The backward pass processes all layers in reverse order, and
a layer's corresponding optimizer step 
can be executed immediately after its backward pass.}
This technique reduces memory footprint because the gradients of those layers can be released \xuehai{right after} their updates are applied.
In addition, this
pipeline is the foundation to achieve
the overlapped execution
of backward pass and optimizer step. 
Note that gradient clipping is often required for training stability~\cite{shoeybi2019megatron}, which necessitates computing the global L2-norm over all gradients and therefore forces the optimizer to wait until the entire backward pass completes.
Recent work breaks this dependency via speculative optimizer steps~\cite{lian2025superoffload}, 
exploiting the observation that gradient clipping rarely modifies gradients in practice.

\textbf{Mixed precision training.}
It accelerates forward and backward computation by leveraging lower‑precision arithmetic~\cite{micikevicius2018mixed}, a common
practice adopted in industrial training pipelines~\cite{shoeybi2019megatron,deepseekai2025deepseekv3technicalreport}.
During the forward pass, inputs and full‑precision parameters (termed \emph{master} parameters) are cast into lower‑precision formats (commonly FP16 or BF16, with recent exploration of FP8 and other quantized variants) as they propagate through the layers.
To preserve numerical stability, selected values such as the loss and certain intermediate activations are retained in full precision.
During the backward pass, gradients are computed in reduced precision, while critical accumulations are promoted to full precision.
Finally, in the optimizer step, gradients are scaled and combined with higher-precision optimizer states (typically FP32, though BF16 has recently been explored) to update the master parameters and optimizer states.

\textbf{Activation checkpointing.}
This memory-saving technique trades additional computation for reduced storage of intermediate activations~\cite{chen2016checkpointing}.
Rather than retaining all activations from the forward pass, it saves only a subset as \emph{checkpoints} and discards the rest.
During backpropagation, before computing gradients for layers between two consecutive checkpoints, the system \emph{recomputes} the missing activations by replaying the forward pass starting from the nearest checkpoint, and then runs the backward pass using the reconstructed activations.
This enables training deeper models or larger batch sizes within the same memory budget.

\textbf{Gradient accumulation.}
Designed to reduce the effective memory footprint of training, it decomposes the computation of a large batch across multiple smaller \emph{micro-batches}
with multiple forward and backward passes.
The gradients produced by all
micro-batches are accumulated in full precision, and
then the optimizer step is applied to update the parameters.
This approach allows training with larger effective batch sizes than would otherwise fit into GPU memory, while preserving the statistical benefits of large‑batch optimization.



\subsection{SSD-Offloaded Training}

\begin{figure}[t]
    \centering
    \includegraphics[width=\linewidth]{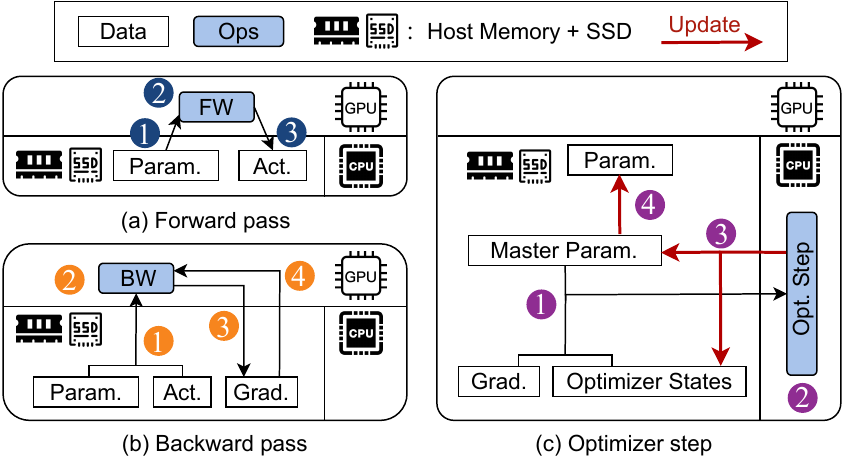}
    \caption{A conceptual diagram of the heterogeneous memory LLM training. Overview of (a) the forward pass, (b) the backward pass, and (c) the optimizer step. (From~\cite{jang2024smart})}
    \label{fig:storage-dataflow}
\end{figure}

Figure~\ref{fig:storage-dataflow} provides an overview of SSD-offloaded LLM training methods with mixed‑precision computation, under the assumption that most training data, including model parameters and optimizer states, must be offloaded.
In this setting, per‑layer activation checkpointing (i.e., checkpointing the input activations of all Transformer blocks) is typically applied for two reasons. First, it reduces activation‑swapping traffic by limiting it to the swapping of only inter‑layer checkpoints, rather than all activations. Second, it ensures that GPU memory only needs to accommodate the footprint of a single Transformer layer’s backward pass at a time, thereby enabling the system to train with larger batch sizes or bigger models.
The detailed data traffic is explained as follows.

In the forward pass (Figure~\ref{fig:storage-dataflow}(a)), the GPU loads the low-precision parameters of an LLM layer (\circled{1}), conducts the layer's forward computation (\circled{2}), and offloads its input activations as the checkpoint (\circled{3}).
\xuehai{The system iteratively performs steps \circled{1}--\circled{3}  over all layers, which can be overlapped via pipelining in practice.}

\xuehai{In the backward pass (Figure~\ref{fig:storage-dataflow}(b)), GPU loads the mixed-precision parameters and checkpointed activations of an LLM layer (\circleo{1}), conducts the layer's backward computation (\circleo{2}), and offloads the resulting gradients (\circleo{3}).}
\yishu{When gradient accumulation is enabled, accumulated gradients are fetched before the upcoming backward computation of the same layer (\circleo{4}) to accumulate new gradients.}
This sequence repeats for all layers in reverse order
and can also be overlapped.

When all forward and backward passes are finished, the system performs optimizer steps (Figure~\ref{fig:storage-dataflow}(c)) for each parameter chunk:
the CPU loads the chunk's gradients, master parameters, and optimizer states (\circlep{1}), conducts the optimizer step (\circlep{2}), and updates the master parameters and optimizer states (\circlep{3}).
The master parameters are then converted into low precision to update the low-precision parameters (\circlep{4}).
This sequence repeats for all parameter chunks, and the forward pass of the next iteration begins after all updates are completed.
The chunk granularity need not align with layer boundaries, since the optimizer step is inherently element-wise.

\xuehai{
Since the master parameters are only involved in optimizer step, we treat them as part of the optimizer states for the remainder of this paper.
Unless otherwise specified, we assume the training optimizer to be Adam~\cite{kingma2017adammethodstochasticoptimization}.
Each model weight element is therefore associated with three full‑precision states, i.e., master parameter, momentum, and variance.
For clarity, {\em ``parameter''} in the subsequent discussion denotes the low‑precision version used in forward and backward passes.
}







%% file: analysis.tex
\section{SSD-Offloaded Training Scheduling}

\subsection{Roofline Model}
\label{sec:roofline}

\begin{figure}[t]
    \centering
    \includegraphics[width=0.65\linewidth]{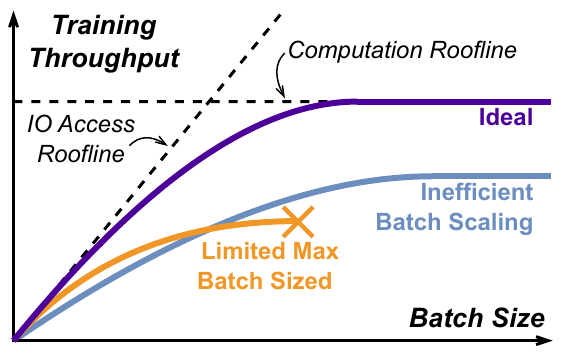}
    \caption{Roofline model of SSD-offloaded Training}
    \label{fig:roofline}
\end{figure}

\xuehai{For simplicity, we assume that 
optimizer states are entirely stored in SSD.
Therefore, each training iteration must load the optimizer states from the SSD once and write them back once, establishing a fundamental I/O bound. }

Figure~\ref{fig:roofline} presents a roofline model for SSD-offloaded training, where the x- and y-axis represents batch size and training throughput. 
Two performance bounds constrain the achievable throughput. First, the \emph{I/O access roofline}---a line passing through the origin---represents the \xuehai{scenario} where iteration time equals the storage access time for optimizer states.
No system can exceed I/O access roofline without reducing per-iteration I/O volume.
Second, the \emph{computation roofline}---a horizontal line---represents the maximum throughput achievable when GPU compute capacity becomes the bottleneck, independent of batch size or memory constraints.

Ideally, an SSD-offloaded training system should first 
exhibit near-linear throughput scaling with batch size, close to the IO access roofline, when optimizer state's I/O access time dominates iteration time, then rapidly converge to the computation roofline.
At this point, the system achieves throughput comparable to training without offloading, effectively hiding all I/O overhead.

However, existing systems exhibit two suboptimal modes.
The first mode suffers from \emph{limited maximum batch size}: architectural constraints prevent sufficient batch size scaling, and the overhead of scaling remains so high that even at maximum batch size, throughput remains far below the computation roofline. 
The second mode, while supporting arbitrarily large batch sizes, demonstrates 
\emph{inefficient batch scaling}: high batch size scaling overhead and an inability to effectively overlap optimizer computation with GPU execution cause throughput to increase slowly and converge prematurely below the computation roofline.


\subsection{Single Forward-Backward Schedule}

A group of SSD-offloaded training systems is explicitly designed and optimized for a single forward-backward pass execution~\cite{fang2022patrickstar,sun2022stronghold,liao2025ratel}.
To boost the training throughput, the common strategy employed is to maximize the training batch size in a single forward-backward pass.
\xuehai{With recomputation, the size of the recovered activations
between checkpoints is bounded by GPU memory capacity.
To accommodate enlarged operation with increased batch size,
checkpoints have to be created more frequently and carefully swapped.}
For example, recent work~\cite{liao2025ratel} additionally checkpoints the activation between the attention layer and the feed-forward network (FFN) inside each Transformer block~\cite{vaswani2017attention}.

\begin{figure}[t]
    \centering
    \includegraphics[width=\linewidth]{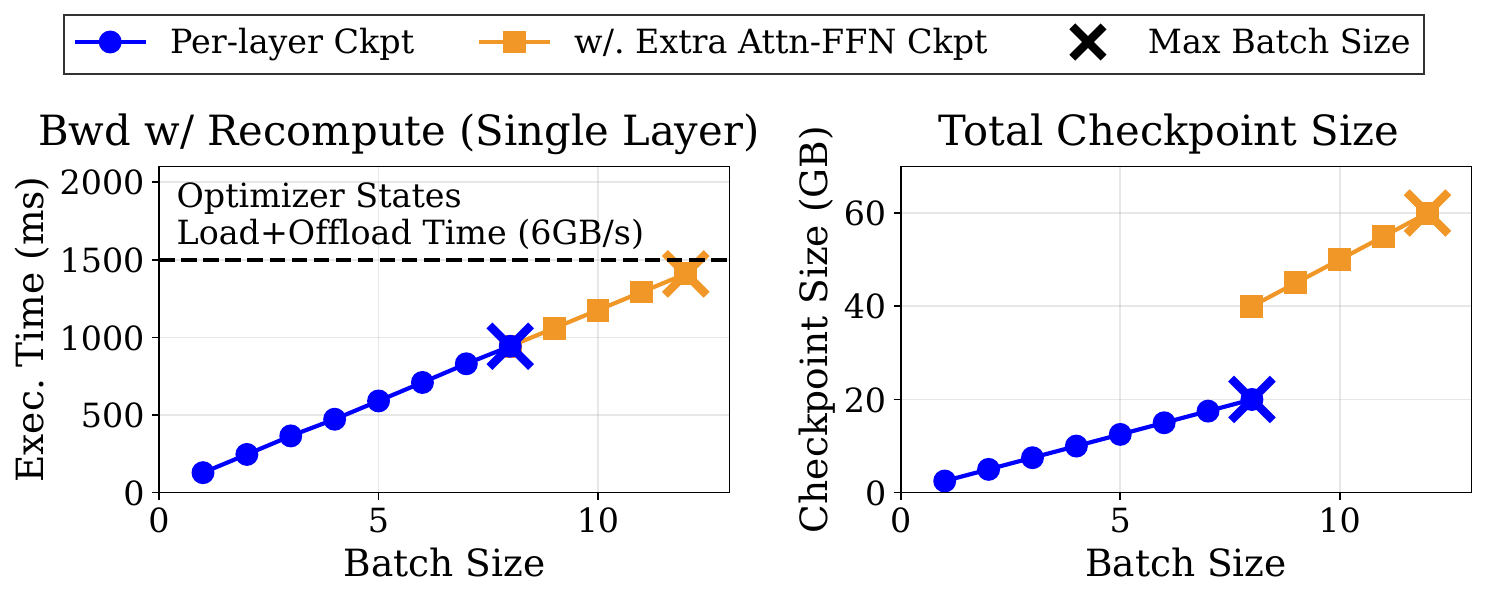}
    \caption{Batch size scaling in single forward-backward schedule. We use GPT-65B (Section~\ref{sec:eval-setup}) as an example.}
    \label{fig:batch-scale}
\end{figure}

As shown by Figure~\ref{fig:batch-scale}, even with such optimizations, the maximum reachable batch sizes are typically not large enough to entirely hide the optimizer step.
From an implementation perspective, 
we admit that by further optimizing such activation checkpointing and swapping~\cite{jain2020checkmate,yuan2024ckpt2,beaumont2021ckpt3},
the maximum achievable batch size may \xuehai{be further increased}.
However, we consider scaling the batch size 
for a single forward-backward pass ineffective for two fundamental
reasons.

First, the most memory-intensive operation in the training computation graph fundamentally caps the maximum achievable batch size.
Second, when increasing the frequency of activation checkpointing to achieve a larger batch size, the total traffic increases superlinearly because not only more 
tensors are swapped, but also the swapped tensors
become larger along the batch dimension. 
Essentially, this approach increases the batch size {\em at an exceedingly high cost}.
As shown in Figure~\ref{fig:batch-scale},
compared to per-layer activation checkpointing, applying extra checkpoints between the attention layer and the feed-forward network can only lead to a 1.5× improvement in the maximum achievable batch size. However, this in turn exacerbates the checkpoint swapping traffic by a factor of 3×, because the extra checkpoints first double the traffic under the same batch size, and the batch size increases by a factor of 1.5×, which jointly leads to the inflation.
Note that an increase in checkpoint size from 20GB to 60GB for one GPU
will be inflated much more drastically when extended to data-parallel training with multiple GPUs.
For example, with 4 GPUs, it will lead to an overwhelming 240GB for checkpoints, 
forcing a higher percentage of---or even almost all---remaining 
training data to be offloaded to SSD,
which consequently leads to higher SSD traffic.


\subsection{Horizontal Gradient Accumulation}
\label{sec:old-schedule}

An alternative approach to scaling the training batch size is to increase the number of forward-backward passes, rather than scaling the batch size 
for a single forward-backward pass. 
This method scales the aggregated batch size by scaling the number of passes with multiple micro-batches.
The computational schedule for gradient accumulation in existing systems, such as ZeRO-series~\cite{rajbhandari2020zero,ren2021zerooffload,rajbhandari2021infinity}, is shown in Figure~\ref{fig:idea}(a).
We refer to this schedule as \emph{horizontal} gradient accumulation, in which the system performs the forward and backward passes of all layers for one micro-batch before moving on to the next.
This solution works well in pure GPU memory training, since for each micro-batch, the activation checkpoints created in the forward pass can be \xuehai{directly} used in the backward pass. It effectively 
ensures a constant peak GPU memory usage that does not scale with the number of micro-batches.

However, we find the horizontal schedule ineffective for SSD-offloaded training for two key reasons.
First, in the horizontal schedule, for each micro-batch, the total low-precision parameters have to be loaded to GPU(s) twice, once during the forward pass and once during the backward pass since recomputation requires the model weights (see Section~\ref{sec:basic}).
\yishu{Additionally,}
for each micro-batch during 
the backward pass,
the buffer for accumulating full-precision gradients must be loaded onto the GPU(s), and written back after accumulation.
These requirements significantly increase the memory traffic between GPU and host CPU memory, and the impact will be even larger if the low-precision parameters or the full-precision gradients have to be offloaded to SSD.

Second, though the batch size can be scaled as needed, the portion of GPU compute that can be overlapped with the optimizer step remains unchanged.
This is because the optimizer step can start only after the last micro-batch completes the backward pass of the last layer and the accumulated parameter gradients have been transferred to CPU memory.
Besides, the common convention~\cite{lian2025superoffload,yuan2025cost,fang2022patrickstar,liao2025ratel} assumes that 
the model is fully updated 
before the start of the next iteration.
As a result, at most the backward pass over all layers of a single micro-batch can be overlapped with the optimizer step, and this overlap does \emph{not} scale even if we increase batch size by using more micro-batches.
\xuehai{The gray box in Figure~\ref{fig:idea}(a) shows the computation that can be overlapped with the optimizer states.}
It lead to the premature saturation of training throughput when training with horizontal gradient accumulation schedule.

\subsection{Proposed: {\em Vertical} Gradient Accumulation}
\label{sec:our-schedule}

We propose {\em vertical} gradient accumulation
scheduling \xuehai{(interchangeable with vertical scheduling)},
a simple yet elegant solution 
to overcome the inefficiencies of horizontal schedule.
Instead of completing all layers of 
one micro-batch before moving onto the next, 
vertical scheduling performs the forward or backward
computation of a specific layer across {\em all micro-batches} before moving to the next layer, 
as shown by Figure~\ref{fig:idea}(b).
This schedule improves parameter reuse and avoids the swapping of gradient accumulation buffers.
Specifically, multiple micro-batches can share the parameters of the same layer, rather than loading them repeatedly.
Similarly, the gradient-accumulation buffer can remain resident in GPU memory while aggregating gradients across all micro-batches, eliminating the need to swap the buffer in and out.

Moreover, vertical schedule also significantly increases
the opportunities to overlap the optimizer step with 
GPU compute.
In particular, with the backward computation finishing
vertically for each layer for all micro-batches,
the optimizer step can be overlapped with the 
backward computation of all micro-batches 
for almost all layers (except 
the last, which is processed first in backward);
and with elaborate implementation, be even
overlapped with part of the forward pass.
\xuehai{Figure~\ref{fig:idea}(b) indicates the clear increase
of overlapping opportunities with vertical schedule.}


It is worth noting that vertical schedule is not a ``free'' lunch. Compared to the horizontal schedule, now the system has to switch between different micro-batches.
This requires the GPU to offload the output activations 
of each micro-batch for every layer 
during the forward pass, 
\xuehai{which will be later needed as the input for 
(1) the next layer in forward pass, and 
(2) the recomputation of the same layer in backward pass. }
Therefore, this schedule essentially trades extra checkpoint swapping for the reduced parameter loading and gradient swapping traffic. 
{\em Is the tradeoff really beneficial?}

\begin{figure}[t]
    \centering
    \includegraphics[width=1\linewidth]{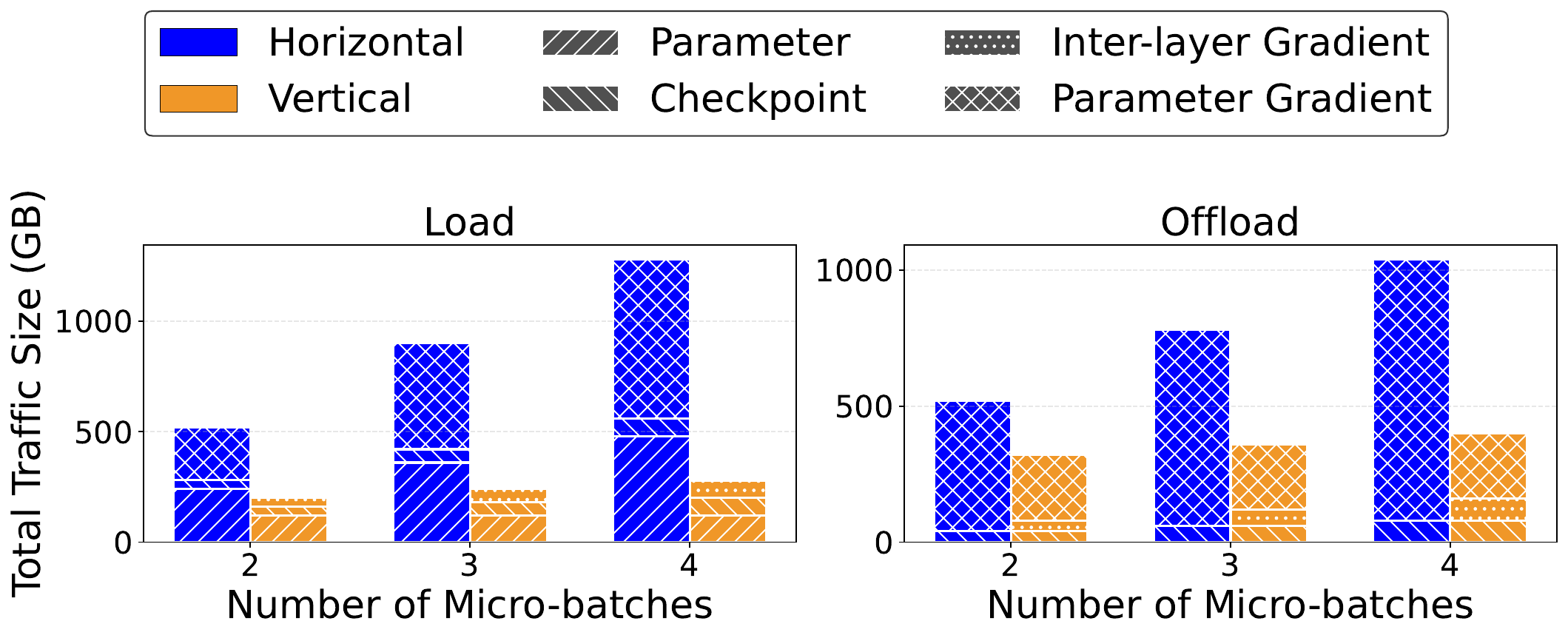}
    \caption{Impact of horizontal vs. vertical scheduling on GPU load and offload traffic. We use GPT-65B as an example.}
    \label{fig:checkpoint_vs_parameter}
\end{figure}

We claim the answer is YES because the two traffic
scales differently with the model's hidden dimension:
(1) the number of elements in each inter-layer activation checkpoint only scales {\em linearly} with the model's hidden dimension; and 
(2) the number of elements in each LLM layer scales {\em quadratically}---mainly due to the projection matrices in FFNs.
Therefore, the size of each activation checkpoint is typically smaller than the size of each LLM layer. 

For an LLM \xuehai{in tens of billions scale or more}, parameter reusing should be prioritized over activation checkpoints. For example, when training a GPT-style 65B model with a micro-batch size equals 8 and sequence length equals 2048 (see Section~\ref{sec:eval-setup}),
the number of elements in each inter-layer activation checkpoints is $8\times2048\times8192\approx1.34\times10^8$ while the number of parameters per LLM layer is approximately $8.05\times10^8$, which is 6× as large.
As shown in Figure~\ref{fig:checkpoint_vs_parameter}, adopting vertical schedule instead of horizontal schedule dramatically reduces both GPU load and offload traffic,
primarily thanks to the reduction of 
the swapping traffic for parameters and gradients 
by a factor close to the number of micro-batches.


%% file: design.tex
\section{Pipelined Vertical Scheduling}
\label{sec:design}

\begin{figure*}[t]
    \centering
    \includegraphics[width=\linewidth]{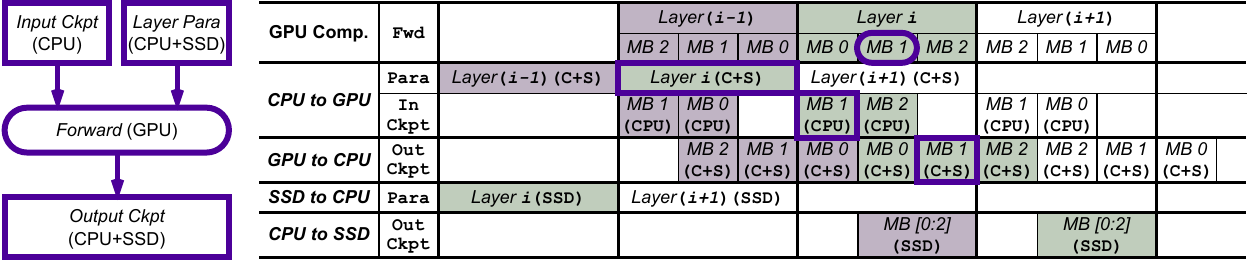}
    \caption{The forward dataflow in GreedySnake (left) and its pipelined version (right).}
    \label{fig:fwd}
\end{figure*}

\begin{figure*}[t]
    \centering
    \includegraphics[width=\linewidth]{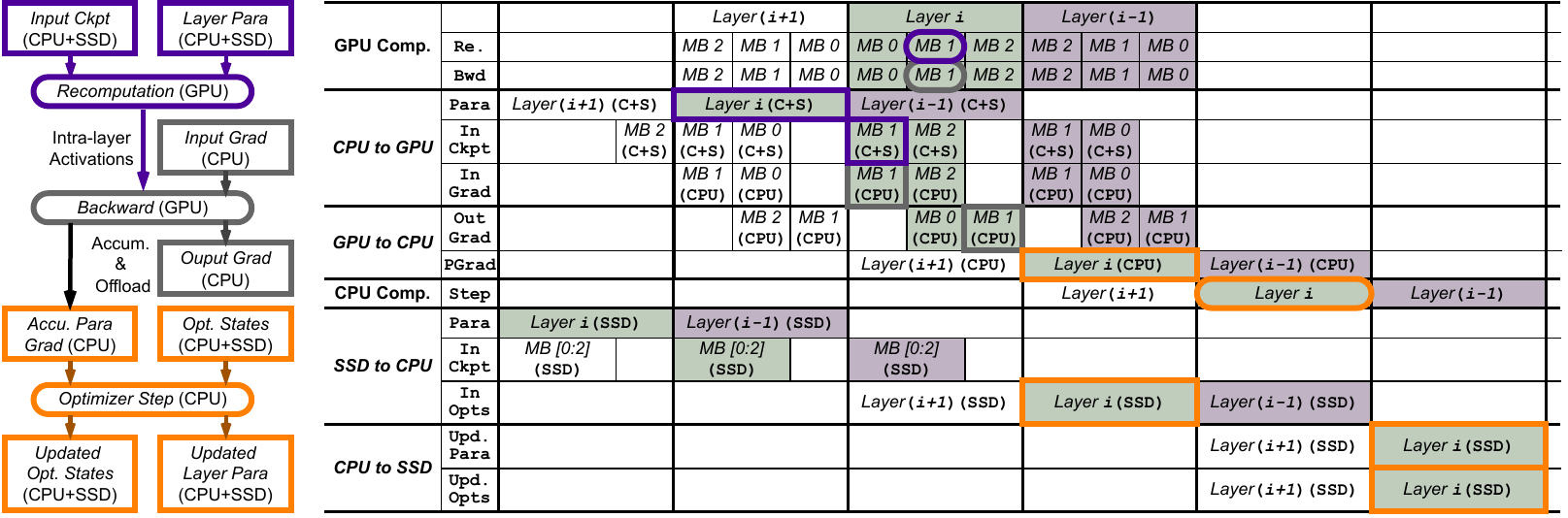}
    \caption{The optimizer-backward overlapping dataflow in GreedySnake (left) and its pipelined version (right).}
    \label{fig:opt+bwd}
\end{figure*}

\begin{figure*}[t]
    \centering
    \includegraphics[width=\linewidth]{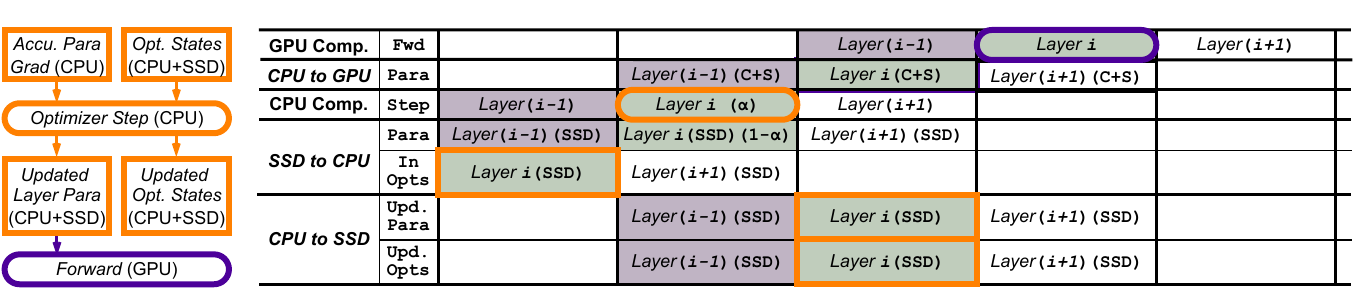}
    \caption{The optimizer-forward overlapping dataflow in GreedySnake (left) and its pipelined version (right).}
    \label{fig:opt+fwd}
\end{figure*}

This section discusses the detailed pipelined
execution of forward pass, backward pass,
and optimizer step with vertical scheduling. 
We focus on two design goals:
(1) excellent utilization of various
system resources, i.e., GPU/CPU computing and
memory, communication bandwidth between 
CPU and GPU, as well as between CPU memory
and SSD; and
(2) efficient overlapping of various 
operations to mitigate the large storage-access overhead and optimize 
performance.

\subsection{Problems and Principles}
\label{sec:problem}

While the idea of vertical scheduling is
intuitive, realizing it efficiently requires
delicate considerations. 
First, we need to determine the granularity of
data transfer to maximize overlapping opportunity.
Second, with the execution of different tasks
being conducted concurrently, we need to 
carefully determine the time to initiate 
the necessary operations, to {\em get
the right data at the right time}.
Third, the system should eliminate the pipeline
bubbles as much as possible. 
To solve these problems,
GreedySnake follows
three {\em design principles}:

\textbf{Differentiated data movement granularity. }
\xuehai{The data traffics between GPU and CPU memory are different 
from those between CPU memory and SSD.}
To ensure efficient storage I/O, the system
performs data transfer between SSD and CPU
memory with larger layer granularity.
\xuehai{
To align with the fact that computation and data fetch in GPU are 
performs at micro-batch granularity, 
data} between CPU and GPU memory is transferred with smaller 
micro-batch granularity, which means that 
the transferred data is divided 
into pieces equal to the number of micro-batches.
Based on this principle, between CPU and GPU
memory, layer parameters are transferred
in micro-batch granularity, 
even if they logically belong to one layer; between 
SSD and CPU memory, activation checkpoints
of all micro-batches for a layer 
are transferred together, even if they
are logically independent.

\textbf{Reversely determined operation scheduling.}
Different kinds of computation, i.e., forward,
backward pass, and optimizer step, 
depend on different inputs stored in 
various locations, i.e., GPU/CPU memory or 
SSD, with different cost of data transfer. 
To ensure the required data is available at the 
right time in the right place, 
we infer the time to initiate data 
transfer reversely starting from the computation where
the data is needed. As shown later, 
this leads to a compact 
and efficient pipeline schedule.

\textbf{Automated configuration space exploration. }
Based on the roofline model in Figure~\ref{fig:roofline}, 
GreedySnake aims to achieve the highest training throughput
with a minimal batch size. 
To achieve the near ideal setting, a huge configuration space,
e.g., the ratios of activation checkpoints, parameters, and optimizer states to be stored in CPU memory and SSD,
needs to be searched, making it impractical to perform
manually. 
We employ a simple yet accurate performance model to predict the forward and backward execution time of a single LLM layer. We then employ linear programming to minimize the 
combined execution time of forward, recomputation, and backward, under 
\xuehai{a given} memory constraint.



\subsection{Forward Pass}
\label{sec:forward}

Figure~\ref{fig:fwd} shows the pipelined execution 
of three consecutive layers with three 
micro-batches for forward pass.
The computation graph shows that, the forward
pass depends on two inputs:
activation checkpoints and layer parameters.
The input checkpoints produced by the previous layer 
are written to the SSD but at the same time cached in CPU memory 
to be consumed by the current layer.
The layer parameters and 
output checkpoints are stored in both SSD and CPU memory.


Each {\em pipeline stage} performs
the forward computation  of a layer after
all parameters are loaded into GPU memory.
Based on the first design principle, 
the checkpoints of micro-batches for this layer
are loaded one by one with a pipeline
within the stage. 
It is illustrated by MB0, MB1, and MB2 
below ``Layer i'' and the overlapped CPU-to-GPU
transfer, e.g., during the execution of MB0, the checkpoint
of MB1 is loaded into GPU. 
Note that the first micro-batch
MB0 does not need to load the 
checkpoint, as long as we ensure consecutive
layers {\em alternate} between opposite
micro-batch execution order, e.g., 
Layer (i-1) follows the order 
MB2 $\rightarrow$ MB1 $\rightarrow$ MB0, and
Layer i follows the order 
MB0 $\rightarrow$ MB1 $\rightarrow$ MB2.
In this way, the checkpoint produced
by MB0 of Layer (i-1) can be kept in GPU memory
and directly used by MB0 of Layer i. 
After a micro-batch, the output checkpoints
are stored to CPU memory and SSD.
For the CPU portion, the data are directly written
to the target memory address; for the SSD portion, 
the data of each micro-batch are first collected in
a buffer in CPU memory, and then written to SSD when all micro-batches
are completed, shown as MB[0:2] (SSD).
This SSD offloading is performed in the 
next pipeline stage, overlapped with the forward pass of 
Layer (i+1).

Next, we consider parameter transfer, which is 
performed in two steps. 
To ensure the availability of parameter in GPU memory
for Layer i's forward pass, the first step---transfer 
parameters in SSD to CPU---is initiated {\em two} pipeline stages 
\xuehai{in advance}. It is the {\em latest} time to schedule
it without introducing \xuehai{wait} for forward pass.
After that, during exactly {\em one} stage before Layer i's forward pass, the second step---transfer parameters from CPU to GPU memory---is scheduled, fully overlapped
with the execution of Layer (i-1)'s forward pass.
\xuehai{Again based on first design principle,
the layer's parameters are transferred in multiple
chunks equal to the number of micro-batches for 
better efficiency.
}

\subsection{Overlap Backward with Optimizer Step}
\label{sec:backard}

Figure~\ref{fig:opt+bwd} shows the pipelined execution of
backward pass of Layer (i-1), Layer i, and Layer (i+1), and how
the optimizer step is performed after Layer i's backward pass.
The computation graph shows the combined
data dependence of backward pass and optimization step.
First, all intra-layer activations are recovered by recomputation 
using input checkpoints produced during forward pass
and layer parameters, both are stored in CPU memory
and SSD. 
The backward pass depends on:
(1) input inter-layer
gradients stored in CPU memory produced by
the previous backward layer; and  
(2) all activations
produced by recomputation.
The output of the backward pass by one micro-batch includes
the output inter-layer gradients passed to the next layer, and 
the parameter gradients accumulated in GPU memory, \xuehai{which are} 
transferred to CPU memory after all micro-batches of a layer are finished. 
The inter-layer gradients are passed between layers
through CPU memory, similar to the inter-layer activation checkpoints in forward pass,
except that they are not written to SSD.

After all micro-batches of a backward layer are
completed,
the CPU optimizer step is executed based on
the \xuehai{fully} accumulated parameter gradients and 
optimizer states partially transferred from SSD.
It generates updated layer parameters and 
optimizer states, both will be stored 
in CPU memory and SSD. 
In the following, we explain how the computation graph 
is realized in GreedySnake.

Consider Layer i, the recomputation is 
divided into micro-batches,
but does require the whole
parameter of the layer. 
Similar to forward pass, they are transferred in two steps
scheduled in two earlier pipeline stages.
Since the input checkpoint can be transferred to GPU memory
in micro-batch granularity, the SSD-to-CPU memory transfer
of this data in layer granularity, i.e., MB[0:2] (SSD), is scheduled in 
{\em one} pipeline stage earlier than Layer i's backward
computation. It is because by the end of that stage, only 
the input checkpoint of MB0 needs to arrive in GPU memory.
That of MB1 and MB2 are transferred within Layer i's backward
computation stage from CPU memory with an in-stage pipeline, 
similar to checkpoints in forward pass. 

After MB-i's recomputation is finished, MB-i's backward
computation \xuehai{depending on the output of 
recomputation and input gradients can start.}
Note that while MB-i of recomputation and backward 
are placed during the same time period, dependence exists
between them, we draw the figure this way to keep it clean
without introducing much confusion. 
The input gradients' behavior is essentially the same as
checkpoints of forward pass between layers: 
Layer i reads them in micro-batch granularity from CPU memory.
Moreover, input gradients are directly
forwarded through GPU memory \xuehai{between the two 
micro-batches before and after layer boundary}. 

Next we consider optimizer step executed on CPU. 
The parameter gradients of Layer i are accumulated 
across all micro-batches in GPU during computation, in the 
next pipeline stage, they are transferred to CPU memory.
In the same stage, the SSD portion of
Layer i's optimizer state is transferred from SSD to CPU
memory.
After both accumulated parameter gradients and
optimizer states are available in CPU memory, the optimizer step
is performed in the following pipeline stage. 
Finally, part of the updated parameters and optimizer states
are written to SSD, in the rightmost pipeline stage. 
All these five key steps are highlighted with orange boxes.

\subsection{Overlap Optimizer Step with Forward} 
\label{sec:opt_fwd}


While vertical scheduling increases the potential of overlapping optimizer step,
moving the optimizer states
between SSD and CPU memory still incurs considerable overhead.
Consequently,
as the number of micro-batches increases and the system shifts from I/O-bound to compute-bound (see Section~\ref{sec:roofline}),
the forward pass becomes compute-bound first,
while the backward pass remains I/O-bound, 
\xuehai{because the I/O-intensive optimizer
step is overlapped with it}.

To improve this obvious suboptimality and spread I/O operations over longer period, 
our main insight is to {\em delay certain percentage of optimizer step to the 
forward pass of the next iteration}. 
\xuehai{We quantify the portion of optimizer step to be delayed
by defining the {\em delay ratio} $\alpha \in [0,1]$.
It also means, for each layer, only $(1-\alpha)$ of optimizer states and parameters are updated in the backward pass.}
As a result, the CPU-SSD traffic of optimizer states by the end of the current iteration 
is reduced by $\alpha$, and the reduced traffic is added to the forward pass of the next iteration.

Delaying $\alpha$ of optimizer step to the next iteration also requires
keeping the accumulated gradients longer, 
requiring extra CPU memory.
However, straightforward designs will likely offset much of the benefits of this optimization:
offloading $\alpha$ of gradients to SSD by itself incurs additional SSD traffic, defeating
the initial purpose; while allocating extra CPU memory for these gradients negatively affects
performance since such memory would have been otherwise used to host offloaded training states. 
In GreedySnake, we made an effort to enable the delayed
optimizer step execution {\em without increasing memory consumption}.

The key insight is to {\em reuse} the CPU memory allocated for offloaded parameters and checkpoints, both of which remain resident throughout training.
\xuehai{Layer by layer during the backward pass, the corresponding
data in such memory becomes obsolete gradually.}
To be specific, the CPU-offloaded parameters 
become obsolete after recomputation, 
\xuehai{after the partial optimizer step, $\alpha$ of them remain
stale and can be reclaimed to store gradients.
Similarly, checkpoints also become obsolete 
after recomputation and the memory can be reused as well.}
Our implementation requires that the gradient memory needed for the delayed optimizer step do not exceed the combined capacity of these two reclaimed memory resources.
After the remaining $\alpha$ fraction of optimizer step completes, the reclaimed parameter buffers are overwritten with the corresponding {\em updated} parameters, bringing all CPU-resident parameters up to date.

Figure~\ref{fig:opt+fwd} shows the pipelined execution of
partially delayed optimizer step and
forward pass. The computation graph is
slightly changed to link the $\alpha$ updated
layer parameters to forward pass of the 
next iteration that depends on them. 
Consider Layer i, $\alpha$ of the optimizer
states used for the delayed execution is 
transferred from SSD to CPU memory,
{\em three} stages 
\xuehai{ahead of} Layer i's forward pass.
In the next stage, the CPU optimizer step is performed, producing the updated $\alpha$ fraction of parameters in CPU memory.
To ensure all updated parameters are available in CPU memory before CPU-to-GPU prefetching, the transfer of the $(1-\alpha)$ fraction of parameters from SSD to CPU is scheduled in the same pipeline stage---two stages \xuehai{ahead of} Layer i's forward pass.
Afterward,
they can be sent together from CPU to 
GPU memory, in the stage right 
before Layer i's forward pass.
Besides, the updated parameters and optimizer states are written back to SSD, bringing all SSD-resident states up to date.
In Figure~\ref{fig:opt+fwd},
the above key steps are highlighted
with orange boxes (part of 
optimizer step) and purple boxes (part of forward pass).

\subsection{LP-based Configuration Search}

We formulate the configuration selection as a small Linear Programming~\cite{Chvatal1983lp} (LP) process.
The overall procedure is summarized
in Algorithm~\ref{algorithm:global_optimizer}.
The goal of the algorithm is to find the smallest micro-batch count $n$ that reaches the saturated training throughput and to record the corresponding step delay ratio $\alpha$ and storage ratios $x$ for system configuration.
As shown, the algorithm first benchmarks the target machine and packs the results into system parameters $\mathcal{M}$. This includes GPU/CPU memory, SSD bandwidth, forward/backward time and other system metrics.
Then, the algorithm searches over micro-batch count $n$ (loop) and step delay ratio $\alpha$ (argmax).
For each $n-\alpha$ pair, the algorithm will build a small LP to find the optimal storage ratio for checkpoints, parameters and optimizer states (gradients are 100\% stored in CPU). Assuming that SSD traffic time and computation can always overlap, we consider their maximum as the effective forward/backward time. Then, the objective function will minimize per-layer iteration time with a regularization penalty to minimize SSD traffic when possible.
\xuehai{While other constraints are considered in developing the algorithm, only three become active at the decision boundary}: CPU memory capacity, GPU computation time, and SSD bandwidth.
With these constraints considered, we ensure that all data fits in the available CPU memory, and that the projected iteration time correctly reflects forward/backward passes overlapped with optimizer steps. By solving this LP, the algorithm identifies the configuration that delivers the highest training throughput for each $(n, \alpha)$ combination. It continues to increase $n$ until throughput stops improving and saturated.
This approach can also estimate the overall iteration time and throughput of the entire model, i.e., per-layer time multiplied by the number of layers, plus time for embedding, LLM heads, and other components.

\setlength{\abovecaptionskip}{0pt}
\setlength{\belowcaptionskip}{0pt}
\setlength{\textfloatsep}{10pt}
\begin{algorithm}[t]
\footnotesize
    \SetNoFillComment
    \DontPrintSemicolon
    \caption{Global Configuration Optimizer}
    \label{algorithm:global_optimizer}
    \SetKwInOut{Input}{Input}
    \SetKwInOut{Output}{Output}
    \SetKwFunction{FCoreSolver}{SolveConfig}
    \SetKwProg{Fn}{Function}{:}{}
    \SetKw{Break}{break}
    \SetKw{Continue}{continue}
    
    \Input{System parameters $\mathcal{M}$}
    \Output{Optimal micro-batch count $n^*$, $\alpha^*$, and storage ratios $\mathbf{x}^*$}
    
    \Fn{\textsc{SolveConfig}($\mathcal{M}$, $n$, $\alpha$)}{
        $\mathbf{x} \in [0,1]^3$ \tcp*{For ckpt, param and opt states} 

        $t_f \gets 
        \max\bigl(\mathcal{M}.\text{fwd\_compute\_time} \times n,\,
                  \text{get\_fwd\_ssd\_time}(\mathcal{M})\bigr)$ 
        $t_b \gets 
        \max\bigl(\mathcal{M}.\text{bwd\_compute\_time} \times n,\,
                  \text{get\_bwd\_ssd\_time}(\mathcal{M})\bigr)$ 
        
        $\text{LP} \gets \text{new LpProblem}$ 
        
        $\text{LP} \mathrel{+}= \text{get\_cpu\_mem}(\mathcal{M}, \mathbf{x})
        \leq \mathcal{M}.\text{usable\_dram}$ 
        $\text{LP} \mathrel{+}= \text{minimize }(t_f + t_b)$ \tcp*{$t_b$ includes recompute time} 
        
        \If{$\text{LP is feasible}$}{
            \Return $(\mathbf{x},\, t_f + t_b)$ \tcp*{storage ratios and iteration time}
        }
        \Else{
            \Return $(\text{None},\, \infty)$
        }
    }
    
    \Fn{\textsc{FindOptimalConfig}($\mathcal{M}$)}{
        $\text{max\_throughput} \gets 0$,\, $\text{best\_config} \gets \text{None}$ \\
        $n \gets 0$,\, $\mathcal{A} \gets \{0.01, 0.02, \dots, 0.50\}$ \\
        
        \While{true}{
            $n \gets n + 1$ \\[2pt]
            
            $\alpha^\star \gets \arg\max_{\alpha \in \mathcal{A}}
            \frac{n}{\FCoreSolver(\mathcal{M}, n, \alpha).\text{iteration\_time}}$ \\
            
            $\text{result}^\star \gets \FCoreSolver(\mathcal{M}, n, \alpha^\star)$ \\
            $\text{throughput}^\star \gets \frac{n}{\text{result}^\star.\text{iteration\_time}}$ \\
            
            \If{$\text{throughput}^\star \geq 1.01 \times \text{max\_throughput}$}{
                $\text{max\_throughput} \gets \text{throughput}^\star$ \\
                $\text{best\_config} \gets \text{result}^\star$ \\
            }
            \Else{
                \Break
            }
        }
        
        \Return $\text{best\_config}$ \\
    }
\end{algorithm}


%% file: implementation.tex
\section{System Implementation}
\label{sec:impl}

GreedySnake is implemented in approximately 5K lines of Python (excluding comments and blank lines) based on PyTorch~\cite{pytorch} to manage computation and tensor storage, with the \texttt{asyncio} and \texttt{cpu\_adam} module reused from ZeRO-Infinity~\cite{rajbhandari2021infinity}. To efficiently scale the computation, ZeRO-style Fully Shared Data Parallellsim~\cite{rajbhandari2020zero,zhao2023fsdp} (FSDP) is integrated. To implement the pipelined execution
based on vertical scheduling discussed in 
Section~\ref{sec:design}, we implement three
coordinators to manage different types of training data.

The first is {\em Inter-layer Tensor Coordinator}, which is responsible
for the data movement of activation
checkpoints in forward pass and inter-layer gradients in backward pass.
The two types of data share similar 
access patterns during execution. 
The second is {\em Parameter Coordinator},
which is responsible for fetching parameters.
The last is {\em Optimizer Step Coordinator} that is responsible for orchestrating
data movements of other data
types, including optimizer states and parameter
gradients. 
The pipelined
execution discussed before
can be considered as a two-dimensional 
resource-time space. 
The coordinators focus on the {\em resource 
dimension}: they together ensure that the correct
set of operations are scheduled for each 
pipeline stage, i.e., a period in
time dimension. 

\begin{figure}[t]
    \centering
    \includegraphics[width=\linewidth]{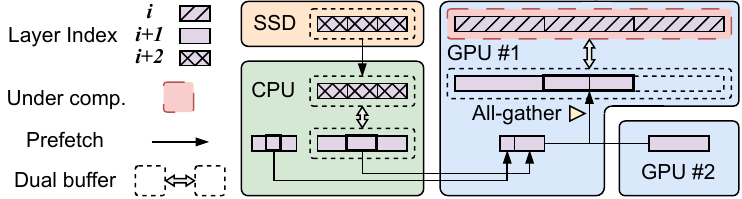}
    \caption{Parameter prefetching in forward pass across the memory hierarchy. We use \#GPU=2, \#MB=3 as an example.}
    \label{fig:impl-para}
\end{figure}

Figure~\ref{fig:impl-para} provides an example
of operations of parameter coordinator
in various components. 
At the given time, it orchestrates
the parameter fetching for {\em three} layers. 
Layer (i+2)'s parameters are fetched in 
layer granularity from SSD to CPU memory; 
Layer (i+1)'s parameters are transferred
from CPU memory to GPU memory in micro-batch
granularity, i.e., the transfer of 
a whole layer is divided into pieces
equal to the number of micro-batches; and 
Layer i's parameters are currently in GPU memory
used by computation.
With two GPUs, 
each piece of parameter in GPU will
be generated by the all-gather operation.
We choose to transfer parameters in
micro-batch granularity for performance reason.
Due to the interference between the checkpoint 
traffic \xuehai{from GPU to CPU} and 
the parameter traffic from CPU to GPU,  
transferring all parameters together prevents
the utilization of the full communication
bandwidth between CPU and GPU. The issue
is largely resolved when the transfer
of different data types are divided
into the {\em same} micro-batch granularity.

Each coordinator allocates various CPU buffers
that should be pinned in CPU memory to 
support efficient asynchronous data movement.
However, as of the recent PyTorch (Version: \texttt{2.9.1+cu128}), individual requests for allocating pinned memory buffer are padded to buffers of power-of-two sizes, leading to 
a waste of at most half of the allocated memory.
GreedySnake typically allocates buffers of the same size, e.g., the CPU checkpoint buffer for all micro-batches and layers. Based on this 
pattern, we use dynamic programming to find 
the set of power-of-two buffers that can hold  a given number of buffers of certain 
size with minimum wasted memory.

%% file: evaluation.tex
\section{Evaluation}

\begin{figure*}[!htbp]
    \centering
    \includegraphics[width=0.8\linewidth]{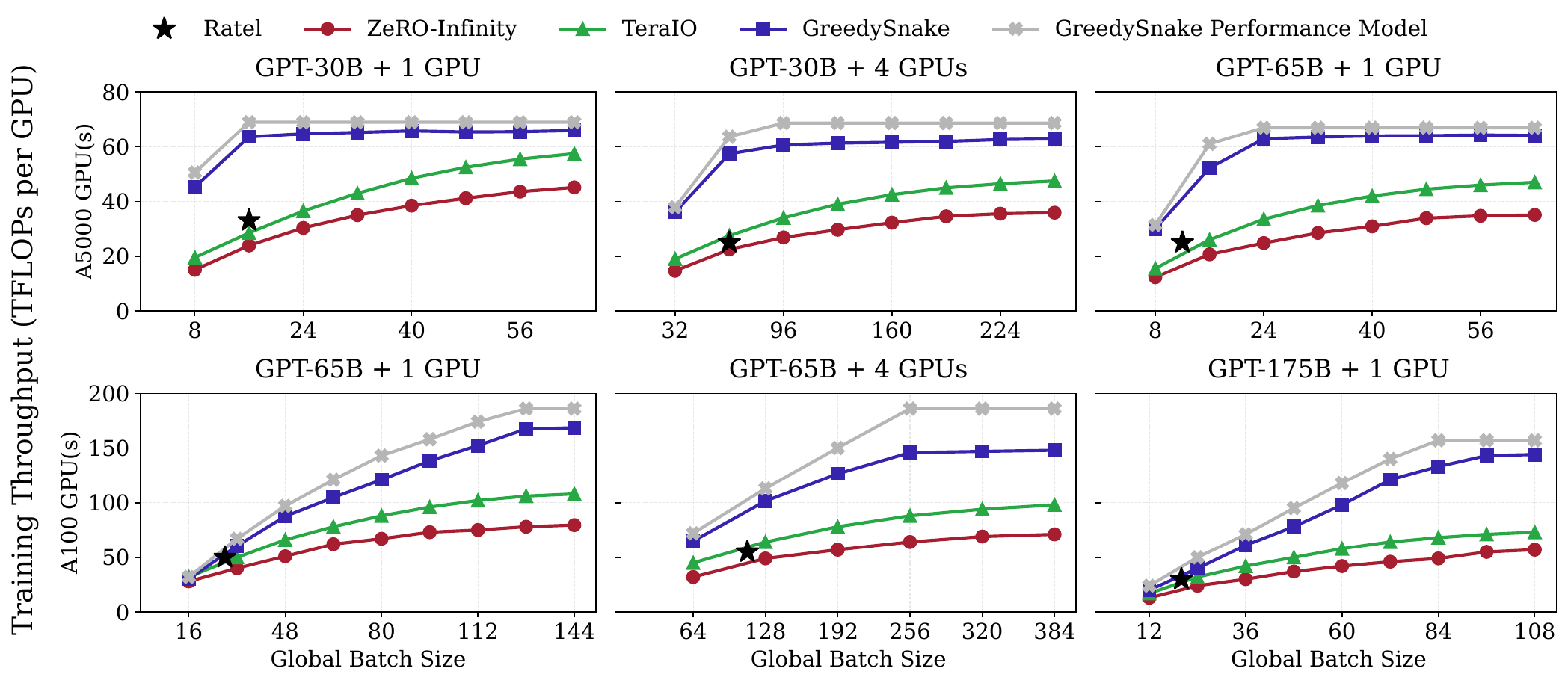}
    \caption{End-to-end throughput of SSD-offloaded training systems when training GPT-style LLMs with sequence length 2048.}
    \label{fig:main}
\end{figure*}

\subsection{Evaluation Setup}
\label{sec:eval-setup}

\begin{table}[t]
	\begin{center}
		\footnotesize
	\caption{Configurations of the evaluation servers}
		\begin{tabular}{|p{1cm}||p{3.2cm}|p{3.2cm}|}
			\hline 
		    & Machine 1 & Machine 2 \\
			\hline
			{\bf CPU} & Dual EPYC 7302 16-Core  & Dual Xeon Platinum 8462Y+ \\
			\hline
			{\bf Mem.} & 256 GB 3200MHz DDR4  & 400 GB 3200MHz DDR4 \\
			\hline
			{\bf PCIe} & PCIe Gen 4  & PCIe Gen 4 \\
			\hline
			{\bf GPU} & NVIDIA A5000 (24GB)  & NVIDIA A100 (40GB) \\
			\hline
			{\bf SSD} & PM9A3 3.84TB NVMe  & 4TB Cloud Storage \\
			\hline
			{\bf CUDA} & V12.1.105   & V12.8.93 \\
			\hline
			{\bf PyTorch} & 2.5.1+cu121  & 2.8.0+cu128 \\
			\hline

\hline			
		\end{tabular}
	\end{center}
	\label{tab:server_config}
\end{table}

\noindent\textbf{Evaluated Machine.} We perform all the experiments on two
servers whose configurations are summarized in Table 1.

\begin{table}[t]
\centering
\footnotesize
\vspace{-0mm}
\caption{Model configuration in evaluation.}
\label{tab:llm_eval}
\begin{tabular}{|c||c|c|c|}
\hline
\textbf{Model Size} & \textbf{\#Layers} & \textbf{\#Heads} & \textbf{Hidden Dimension} \\
\hline
30B & 48 & 56 & 7168 \\
\hline
65B & 80 & 64 & 8192 \\
\hline
175B & 96 & 96 & 12288 \\
\hline
\end{tabular}
\end{table}

\noindent\textbf{Workloads.} Same as related work~\cite{liao2025ratel,fang2022patrickstar,rajbhandari2021infinity}, we use GPT-style LLMs provided by Megatron-LM~\cite{shoeybi2019megatron}. The model configuration is shown in Table~\ref{tab:llm_eval}.

\noindent\textbf{Baseline Configurations.}
We compare GreedySnake against the following open-sourced SSD-offloaded training frameworks:
(1) \emph{ZeRO-Infinity}~\cite{rajbhandari2021infinity}: A widely deployed, production-grade SSD-offloaded training framework that optimizes memory management through coordinated prefetching and computation-communication overlap. We enable activation checkpointing with CPU memory offloading to maximize per-pass batch size. Parameters and optimizer states are offloaded to SSD by default, while parameters are retained in CPU memory when capacity permits to reduce I/O overhead.
(2) \emph{Ratel}~\cite{liao2025ratel}: An SSD‑offloaded training system that co‑designs fine‑grained activation checkpointing with checkpoint swapping across the memory hierarchy. We reuse the original open-source codebase for single-GPU training; for multi-GPU training, we extend it with FSDP on a best-effort basis.
(3) \emph{TeraIO}~\cite{yuan2025cost}: An SSD-offloaded training system that optimizes data movement through tensor lifetime analysis. Because the open-source codebase does not support \xuehai{activation checkpointing and is coupled with TorchTitan}~\cite{liang2024torchtitan}, we use its tensor lifetime profiler to analyze ZeRO-Infinity execution traces and report the throughput projected by the framework's analyzer under the optimized tensor offloading and prefetching plan.
This baseline represents a highly optimized, horizontally scheduled gradient-accumulation scheme with effective tensor management.

\subsection{End-to-End Throughput Comparison}

Figure~\ref{fig:main} shows the end-to-end throughput with different global
batch sizes of GreedySnake
compared to the other systems.
In the smaller A5000-cluster, we train GPT-30B with 1 and 4 GPUs, and 
GPT-65B with 1 GPU. 
In the larger A100-cluster, we follow a similar practice but with larger models:
training GPT-65B with 1 and 4 GPUs and GPT-175B with 1 GPU.
Both ZeRO-Infinity and Ratel fail to support the 4×A5000-65B and 4×A100-175B configurations due to severe CPU memory fragmentation. In contrast, GreedySnake successfully trains under these settings, so we omit the corresponding figures. We report that, on GreedySnake, training GPT‑65B and GPT‑175B with 4 GPUs across the two clusters achieves 63.1 and 128.3 TFLOPs per GPU, respectively.

For both ZeRO-Infinity and GreedySnake,
we choose the largest possible micro-batch size the system can support. 
It is the most favorable choice
for ZeRO-Infinity: as the representative system using gradient accumulation with horizontal scheduling,
larger micro-batch size is better because it can reduce the parameter loading \yishu{cost}.
The global batch size is generated by increasing
the number of micro-batches based on the chosen size for ZeRO-Infinity.
For example, for GPT-30B on the A5000-node, the largest batch size for 
ZeRO-Infinity is 8, and we use 8, 16, 24, ..., 64 as the global batch size.
We choose the largest global batch size when GreedySnake's throughput is
saturated and well beyond the shifting point from I/O-bound to compute-bound.

For GreedySnake, the micro-batch size is much smaller, typically \xuehai{1 or 2},
because the system allocates more memory to support the pipeline, e.g.,
two copies of the gradients are required in GPU memory
for ``vertical'' accumulation across micro-batches.
Based on the micro-batch size, the percentage of data distribution 
among SSD and CPU memory, and the delay ratio are determined by 
LP-based configuration search.
\xuehai{Since LLM training is compute-bound, even the smaller
micro-batch size can well utilize the compute resource}.
In this sense, GreedySnake offers a new way to devote the memory resource 
to support efficient pipelined execution: using small but enough micro-batch sizes.
In contrast, ZeRO-Infinity prevents this possibility since small micro-batch size
will directly lead to a significant increase of traffic for moving parameters. 
Compared to ZeRO-Infinity, the result convincingly shows that GreedySnake achieves
much higher throughput with much smaller global batch size, indicating
a significant advance over the state-of-the-art. 
\yikang{Overall, compared to ZeRO-Infinity, GreedySnake improves saturated throughput by $1.96\times$ and $1.93\times$ on 1 and 4 A100 GPUs for GPT-65B, and by $2.53\times$ on 1 A100 GPU for GPT-175B.}

Among the other systems, Ratel exemplifies scaling batch size within a single forward-backward schedule.
For each configuration, we report only the largest batch size that the system can support.
As discussed earlier, this approach fails to scale the global batch size, and the resulting throughput remains well below saturation.
These results provide clear evidence that the single forward-backward regime is fundamentally unsustainable.
At the same global batch size, we observe that Ratel performs slightly better than ZeRO-Infinity because Ratel (1) overlaps the backward pass with the optimizer step, and (2) prefetches parameters layer by layer, whereas ZeRO-Infinity does not assume the model consists of uniform layers.
The second factor enables a more uniform and compact prefetching pipeline in Ratel.
We also find that Ratel's advantage over ZeRO-Infinity diminishes with 4 GPUs, because checkpoint sizes increase with data parallelism, forcing the system to devote substantially more CPU memory to store checkpoints.

\xuehai{TeraIO focuses 
on analyzing
the lifetime of each tensor
to generate better offloading and prefetching 
plan based on the tensor access pattern 
obtained with profiling.
We apply this methodology to the 
tensor access trace of ZeRO-Infinity, and 
observe that the throughput of
the optimized execution is higher than ZeRO-Infinity
and scales slightly better with global batch size.}
On the one hand, the result shows the potential for 
tensor lifetime analysis. On the other hand, 
the smaller improvements over ZeRO-Infinity
compared to the larger gap from GreedySnake
indicate the limitations of the ``local'' optimization:
without changing the inefficient horizontal global 
scheduling, the potential of throughput improvement
is relatively small. In fact, this result indirectly 
confirms the importance of vertical scheduling. 

Finally, we compare the throughput of GreedySnake with
the estimation from performance model. 
In most cases, the gap is quite 
small, caused by the unavoidable 
dynamic behavior such as pipeline bubble, and 
imbalanced operation time due to the resource 
contention that cannot be predicted accurately.
We also observe that the gap can be slightly larger
when the execution transitions from I/O-bound
to compute-bound. This is because the pipeline
around this period is more sensitive to the variability
of the latency of individual operations. 
\yishu{Additionally, we find that} for GPT-65B with 4 GPUs, the gap is 
larger compared to others. This is because the
cloud cluster is shared by multiple users, \yishu{and as a result,} the SSD \yikang{bandwidth} tends to diverge more from the 
performance model due to the sharing and contention 
of its bandwidth.

\subsection{Benefits of Delayed Optimizer Step}

\begin{figure}[htbp]
    \centering
    \includegraphics[width=\linewidth]{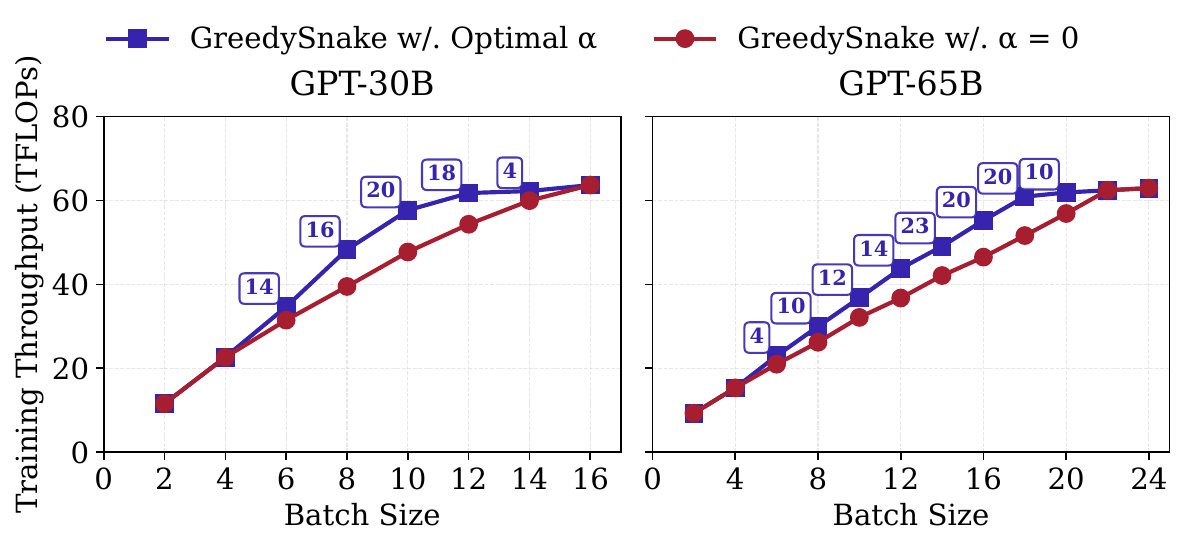}
    \caption{Training throughput w/. and w/o delaying optimizer step. Delaying factors $\alpha$ are annotated in percentage.}
    \label{fig:delay}
\end{figure}

Figure~\ref{fig:delay} compares the throughput of
GreedySnake with and without delayed optimizer step.
For the latter, the delay ratio $\alpha$ is 0.
We see that both cases eventually
reach the same saturated throughput, but the delayed
optimizer step reduces the batch size to reach
that throughput. 
In other words, it makes the I/O-bound phase closer
to the ideal roofline, consistent with the 
fact that this mechanism mitigates the I/O bottleneck
of optimizer step by extending its allowable
execution period.


\subsection{Benefits of Vertical Scheduling}

\begin{figure}[htbp]
    \centering
    \includegraphics[width=\linewidth]{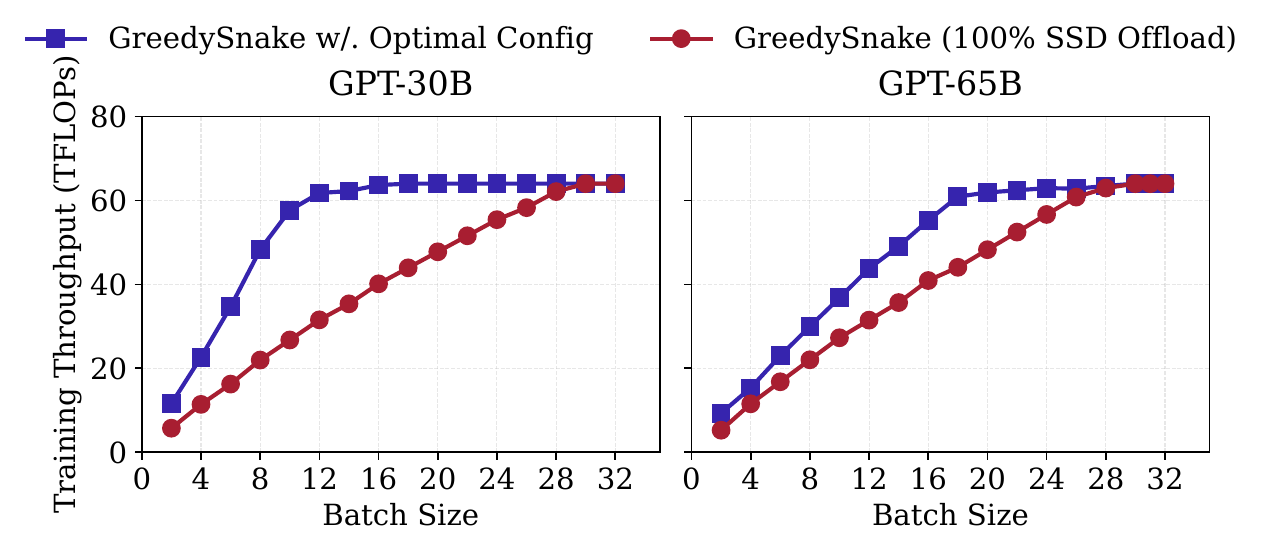}
    \caption{Training throughput achieved by 100\% SSD-offloading vs optmial config.}
    \label{fig:fundamental}
\end{figure}

In this section, we address a key question: is the throughput improvement driven by caching more training data in CPU memory, or by the \emph{fundamental merits} of vertical scheduling?
To answer this, we run an experiment under an extreme setting that forces all training data to be fully offloaded to SSD (i.e., CPU memory is used only for working buffers).
As shown in Figure~\ref{fig:fundamental}, compared to the earlier results using the best configuration found by our LP-based search, throughput increases more slowly.
Crucially, even in this extreme setting, GreedySnake eventually reaches a \emph{similar} saturated throughput.
These results provide the strongest evidence that the intrinsic properties of vertical scheduling are the primary reason for the throughput improvement.

The \emph{crux} of the advantage is that, when GreedySnake adds a new micro-batch to increase the global batch size, the additional computation time---the extra forward and backward work---is \emph{greater than} the additional I/O time to move that micro-batch's activation checkpoints.
Consequently, each additional micro-batch yields \emph{time ``credit''} that can be used to overlap the optimizer step.
\xuehai{For GPT-65B, for example, the forward and backward computation of one micro-batch takes 16.4\,s, whereas the additional checkpoint I/O incurred by that micro-batch takes only 1.1\,s.}
Because the former is much larger than the latter, GreedySnake can still reach---albeit more slowly---a similar saturated throughput even when no training data are allowed to remain in CPU memory: the time ``credit'' accumulates across micro-batches and eventually overlaps the entire optimizer step.
From this perspective, we can unify the benefits of delaying the optimizer step and placing a fraction of training data in CPU memory: both are engineering techniques to \emph{increase the time ``credit''} contributed by each micro-batch, enabling the optimizer step to be overlapped with \emph{fewer} micro-batches.
However, they are not the determining factors for achieving a higher saturated throughput; \emph{the vertical scheduling is}.

\subsection{Training Accuracy}

\begin{figure}[htbp]
    \centering
    \includegraphics[width=\linewidth]{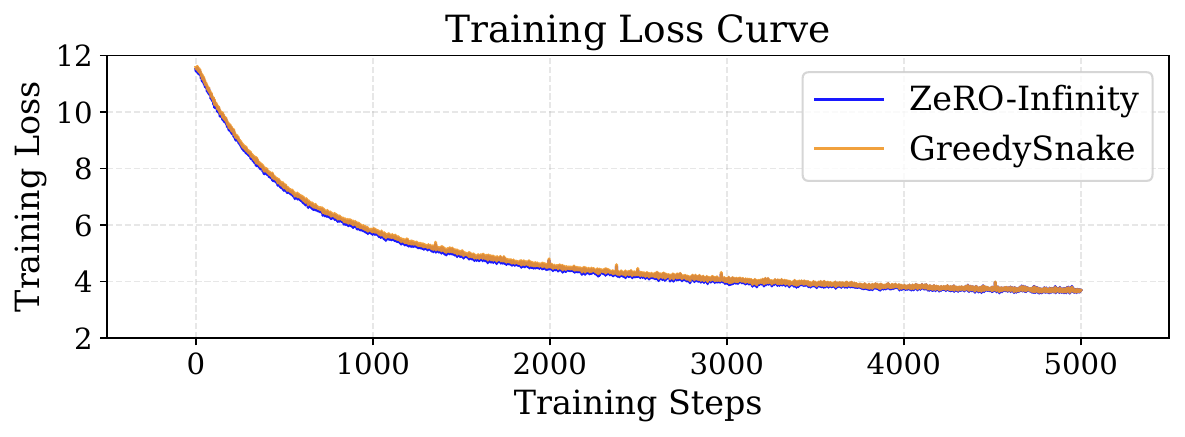}
    \caption{Training loss of GPT‑30B on the Pile dataset.}
    \label{fig:loss}
\end{figure}

Figure~\ref{fig:loss} compares the training loss curves of ZeRO-Infinity and GreedySnake. Both systems produce similar training loss despite minor discrepancies. After ruling out non-determinism from FlashAttention's backward computation~\cite{dao2024flashattention}, we identify two causes. First, different scheduling methods lead to different dropout operation orders, which changes the loss. Second, in ZeRO-Infinity, when the optimizer computation size is not an exact multiple of the CPU's SIMD width, the remaining elements are handled with scalar operations. In contrast, GreedySnake performs all computations with SIMD operations to ensure reproducible loss values across different partition ratios. These implementation differences account for the minor discrepancies observed.


